\documentclass[sigconf]{acmart}

\usepackage{balance}
\usepackage{algorithmic}
\usepackage{algorithm}
\usepackage{multirow}

\def\BibTeX{{\rm B\kern-.05em{\sc i\kern-.025em b}\kern-.08emT\kern-.1667em\lower.7ex\hbox{E}\kern-.125emX}}
    
\setcopyright{acmcopyright}
\copyrightyear{2019}
\acmYear{2019}
\acmConference[CIKM '19]{The 28th ACM International Conference on Information and
Knowledge Management}{November 3--7, 2019}{Beijing, China}
\acmBooktitle{The 28th ACM International Conference on Information and Knowledge
Management (CIKM '19), November 3--7, 2019, Beijing, China}
\acmPrice{15.00}
\acmDOI{10.1145/3357384.3357981}
\acmISBN{978-1-4503-6976-3/19/11}

\begin{document}

\fancyhead{}
\title{Multi-stage Deep Classifier Cascades for Open World Recognition}

\author{Xiaojie Guo}
\affiliation{
  \institution{George Mason University	}
  \streetaddress{4400 University. Dr.}
  \city{Fairfax}
  \state{VA}
  \postcode{22030}}
\email{xguo7@gmu.edu}

\author{Amir Alipour-Fanid}
\affiliation{
  \institution{George Mason University	}
  \streetaddress{4400 University. Dr.}
  \city{Fairfax}
  \state{VA}
  \postcode{22030}}
\email{aalipour@gmu.edu}

\author{Lingfei Wu}
\affiliation{%
  \institution{IBM Research AI}
  \city{New York}
  \postcode{10598}}
\email{wuli@us.ibm.com}

\author{Hemant Purohit}
\affiliation{
  \institution{George Mason University}
  \streetaddress{4400 University. Dr.}
  \city{Fairfax}
  \state{VA}
  \postcode{22030}}
\email{hpurohit@gmu.edu}
 
\author{Xiang Chen}
\affiliation{
  \institution{George Mason University}
  \streetaddress{4400 University. Dr.}
  \city{Fairfax}
  \state{VA}
  \postcode{22030}}
\email{xchen26@gmu.edu}

\author{Kai Zeng}
\affiliation{
  \institution{George Mason University	}
  \streetaddress{4400 University. Dr.}
  \city{Fairfax}
  \state{VA}
  \postcode{22030}}
\email{kzeng2@gmu.edu}

\author{Liang Zhao}
\affiliation{
  \institution{George Mason University	}
  \streetaddress{4400 University. Dr.}
  \city{Fairfax}
  \state{VA}
  \postcode{22030}}
\email{lzhao9@gmu.edu}

%

%
\begin{abstract}
At present, object recognition studies are mostly conducted in a closed lab setting with classes in test phase typically in training phase. However, real-world problem are far more challenging because: i)~new classes unseen in the training phase can appear when predicting; ii)~discriminative features need to evolve when new classes emerge in real time; and iii)~instances in new classes may not follow the ``independent and identically distributed" (iid) assumption. Most existing work only aims to detect the unknown classes and is incapable of continuing to learn newer classes. Although a few methods consider both detecting and including new classes, all are based on the predefined handcrafted features that cannot evolve and are out-of-date for characterizing emerging classes. Thus, to address the above challenges, we propose a novel generic end-to-end framework consisting of a dynamic cascade of classifiers that incrementally learn their dynamic and inherent features. The proposed method injects dynamic elements into the system by detecting instances from unknown classes, while at the same time incrementally updating the model to include the new classes. The resulting cascade tree grows by adding a new leaf node classifier once a new class is detected, and the discriminative features are updated via an end-to-end learning strategy. Experiments on two real-world datasets demonstrate that our proposed method outperforms existing state-of-the-art methods.
\end{abstract}

%
%

%
\maketitle

\section{Introduction}
\begin{figure}[htb]
\centering
\includegraphics[width=.86\linewidth]{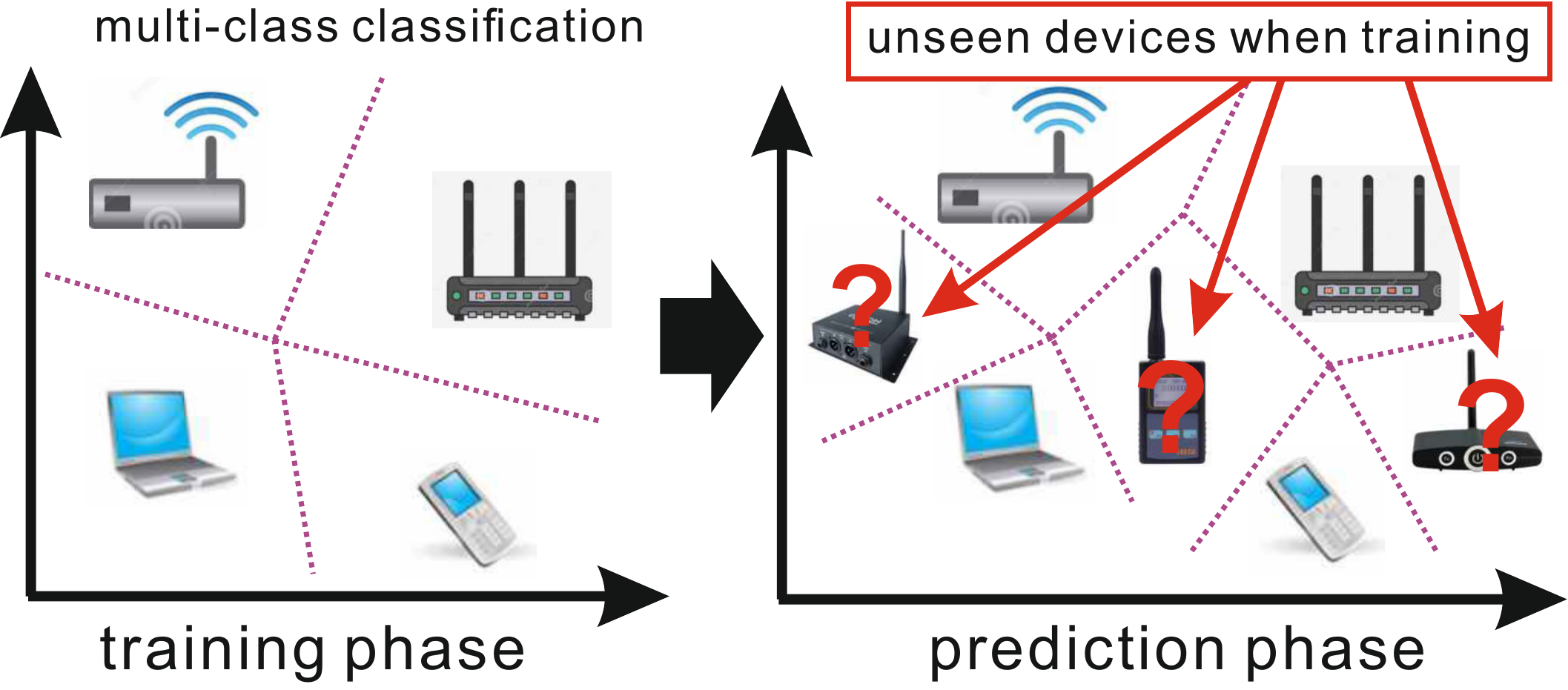}
\caption{In recognizing the device type through the emitted signal fingerprinting, going beyond
conventional multi-class recognition on four known devices, three unseen/new devices must be continuously identified and
distinguished in the prediction phase.}
\label{fig1}
\end{figure}

In recent decades, the development of machine learning and pattern recognition techniques has enabled many different kinds of objects to be learned, detected and analyzed for practical applications~\cite{parkhi2015deep,thrower2000recognition,lei2017similarity,zhao2018distant,guo2016hierarchical,zhao2018prediction}. However, although most recognition systems are designed for a static closed condition, assuming that all the classes are known in the training phrase~\cite{bendale2016towards,wu2018random}, the real world is an \textbf{open set environment}, where only a small part of the entire set of objective classes is known. This requires the recognizer to be able to reject unknown/unseen classes that appear during the inference stage, which has been addressed by several researchers~\cite{nicolau2016hybrid,bendale2016towards}. A real-world application should not only be able to infer but also to classify these newly emerging classes. For instance, in a device recognition problem using device signal fingerprinting in the area of cyber-security, the abundant device types make it impossible to include all the possible data types in the training phase. The example in Fig.~\ref{fig1} demonstrates what happens when there may be four known types of devices in the training phase and three new types that are not seen in the training phase but are present in the prediction phase. Here, the recognizer must not only be able to distinguish the four known devices and be aware of the existence of new types, but must also learn to classify these unknown devices during test phase. To address the above task, the \textbf{Open World Recognition} (OWR) (which is also named as Open World Learning) problem has been formalized. This is composed of three tasks: the classification of known classes, the  detection of unknown classes, and updating the model to learn new classes in an environment where new classes come continuously~\cite{bendale2015towards}. 

As a new domain, OWR is starting to attract increasing attention, with the limited number of existing works being divided into two categories. The first category assumes that a batch of instances from new classes emerge at about the same time. For instance, the time series being divided into several instances can appear together. Two metric-based learning models have been proposed to detect unknown classes and include new classes using the Nearest Class Mean~\cite{bendale2015towards,xu2018learning}. Another category assumes that instances of new classes come continuously, and that the misclassified instances may then have a critical impact on the subsequent learning. Thus, an incremental random tree~\cite{mu2017classification} with a buffer to store the emerging instances from new classes has been proposed; an alternative approach that has been suggested is based on a heuristic recognition model with side-information for the emerging classes~\cite{lonij2017open}. Both these categories require side-information about which instances are from the same classes (e.g., for Twitter topic recognition, tweets with the same hashtags are from the same topics). However, all the methods proposed so far for open-world recognition are highly dependent on pre-defined handcrafted features and is consequently difficult to learn unique and inherent features in real-time as new classes arise. Furthermore, handcrafted features do not necessarily include discriminative and critical features. For example, the device type recognition problem based on signal fingerprinting in Fig.~\ref{fig1} is very difficult since the contents carried in the signals are varied and mixed with ambient noise. This makes it hard to discover any distinguishable features buried in the raw signals purely based on human domain knowledge.

Currently, the end-to-end OWR problem, where the features of new classes need to be automatically extracted and included, cannot be handled by the existing techniques due to the following challenges: 1)~\textbf{It is difficult to update features for an end-to-end model whose architecture is predefined}. The involvement of new classes incurs the need to learn more and new features. However, the architectures proposed in existing studies are predefined and cannot be updated to include new classes. Retraining a model with an expanded architecture to update its features is time-consuming and may result in catastrophic forgetting problems~\cite{goodfellow2013empirical}, where the already learned features are replaced by the newly aquired ones. 2)~\textbf{Features learned based on existing classes are not sufficient to detect future unknown classes}. Newly updated features should also be able to detect future unknown classes whose instances are not available when updating the current model. Even though we assume that the features can be updated with the new classes, they may still be out-of-date when it comes to distinguishing future potential new classes. 3)~\textbf{It is difficult to utilize the relationships among instances from the same class}. Since instances in the same classes may not be independent and identically distributed, enforcing the feature similarities is critical but challenging for end-to-end supervised learning, which is commonly conducted without considering the relationships among instances. 

To address the above challenges and explore the open world recognition problem based on end-to-end learning, we propose a novel Multi-stage Deep Classifier Cascades (MDCC) architecture that leverages a generic cascade architecture to incrementally learn the new class in an end-to-end fashion without retraining the whole model. A one-class classifier that is trained based on a dynamic reference set is proposed to include new classes as well as detect potential future unknown classes to address the second of the above challenges. A self-describing regularization is also proposed to utilize the relationships among instances to address the third of the above challenge. The proposed MDCC has the following distinct features.\vspace{-0.1cm}

\begin{itemize}
\item \textbf{The development of a new end-to-end framework for open world recognition}. To the best of our knowledge, the proposed MDCC is the first generic framework to address the open world problem based on automatic feature extraction utilizing a novel classifier cascade for identifying multiple classes.
\item \textbf{The proposal of a one-class classifier based on a dynamic reference set}. One-class classifiers are proposed and leveraged as leaf nodes that not only automatically learn new features but also detect any potential future unknown classes. Using the proposed dynamic reference set to train the one-class classifier ensures the scalability of the computation and the descriptiveness of the extracted feature.
\item \textbf{The proposal of a self-describing regularization that utilizes the relations among the instances}. In order to enhance the  learning of the features of the new classes in the leaf nodes and narrow the feature spaces of the new classes, a new regularization is proposed by minimizing the inconsistency of features among the related instances.
\item \textbf{Extensive experiments to validate the effectiveness of the proposed model}. Extensive experiments on two real-world datasets demonstrates that MDCC is capable of incrementally detecting and learning emerging new classes, significantly outperforming comparison methods.
\end{itemize}

\section{Related Work}
\textbf{Open set learning}: Open set recognition was first introduced in~ \cite{scheirer2013toward}, which considers the problem of detecting unseen classes that are never seen in the training phase~\cite{zhang2005probabilistic,markou2003novelty}. Many open-set recognition methods based on SVM~\cite{mensink2013distance,bodesheim2015local} and NCM~\cite{bendale2016towards} have since been proposed, but all built on shallow models for classification. Scheirer et al.~\cite{scheirer2013toward} formulated the problem of open set recognition for static one-vs-all learning scenario by balancing open space risk while minimizing empirical error,going on to extend the work to multi-class
settings by introducing a compact abating probability model~\cite{scheirer2014probability}. For the scalability problem, Fragoso et al.~\cite{fragoso2013evsac} proposed the use of a scalable Weibull based calibration for hypothesis generation to model matching scores, but did not address its use for the general recognition problem. Bendale et al.~\cite{bendale2016towards} proposed a novel detection method dealing with deep model architecture by introducing an openmax layer, while Perera et al.~\cite{perera2018learning} proposed a one class classification based on the DCNN which can be used as a novel detector and outlier detector for a single known class. However, none have not addressed the problem of how to incrementally update their model after a new class has been recognized.

\textbf{Incremental learning:} Incremental learning refers to a continuous learning process with new data that has been labeled~\cite{lin2011large,rebuffi2017icarl,zhang2018robust,wang2018incomplete}. Many incremental methods are based on SVM~\cite{lin2011large,cauwenberghs2001incremental} and random tree methods~\cite{rebuffi2017icarl}. Cauwenberghs
et al.~\cite{cauwenberghs2001incremental} proposed an incremental binary SVM by means of saving and updating KKT conditions, while Yeh et al.~\cite{yeh2008dynamic} extended
this approach to include object recognition and demonstrated
multi-class incremental learning. However, incremental SVMs suffer from multiple drawbacks. The update process is extremely expensive (quadratic in the number of training examples learned) and depends heavily on the
number of support vectors~\cite{laskov2006incremental}, so Rebuffi et al.~\cite{rebuffi2017icarl} proposed a memory-controlled training strategy that learns the coming classes as well as maintains the training load. Goodfellow et al.~\cite{goodfellow2013empirical} addressed the knowledge transfer concept in neural network incremental learning. Inspired by this, an error-driven based convolution neural network was introduced by Xiao et al. ~\cite{xiao2014error}, whose model is extended like a tree structure to avoid the catastrophic forgetting problem. Rusu et al.~\cite{rusu2016progressive} proposed a progressive network that retains a pool of pretrained models in training and then learns lateral connections from these to extract features for new tasks. However, all these multi-class incremental learning methods and incremental classifiers are incremental only in terms of additional training samples, not additional training categories. Thus the representation drift of incremental learning is not a problem considered in our problem, where each class has a true and static description over time~\citep{bendale2015towards,de2016online}.

\textbf{Scalable Learning}: different from incremental learning problem, other researchers have proposed tree based classification methods to address the scalability of object categories in large scale visual recognition challenges~\cite{marszalek2008constructing,liu2013probabilistic,deng2011fast,everingham2010pascal}. Recent advances in the deep learning domain\cite{krizhevsky2012imagenet,simonyan2014very} of scalable learning have resulted in state of the art performances, which are extremely useful when the goal is to maximize classification/recognition performance. These systems assume a priori availability of comprehensive training data containing both images and categories. However, adapting such methods to a dynamic learning scenario becomes extremely challenging. Adding object categories requires retraining the entire system, which could be unfeasible for many applications. As a result, these methods are scalable but not incremental.

\textbf{Open world recognition:} Open world recognition considers both detection and learning to distinguish the new classes. Bendale et al. proposed a NCM learning algorithm that relies on the estimation of a determined threshold in conjunction with the threshold counts on some known new classes~\cite{bendale2015towards}. For a more practical situation, Rosa et al. proposed an online-learning approach that involves the NBC classifier instead of NCM~\cite{de2016online}, while Mu et al~\cite{mu2017classification} proposed an online learning for streaming data where new classes come continuously. It is worth noting that Bayesian non-parametric models~\cite{akova2012self,dundar2012bayesian} are not related to our problem. Though they were originally proposed to identify mixed components or clusters in the test data that may cover unseen classes, their clusters are not themselves classes and multiple clusters must be mapped to one class manually.

\section{Problem Formulation}
\label{section:problem}
This paper focuses on the new problem of end-to-end open world recognition (OWR), where instances with various classes arrive continuously and must be recognized. Once a new class is detected, the multi-class recognition model should incrementally learn to accept/classify instances with this class. The following argument provides the notations and the mathematical problem formulation.

\begin{table}
    \centering
    \caption{Important notations and descriptions}
    \begin{tabular}{ll}
    \hline\hline
         Notations &Descriptions \\
         \midrule
         $X_n$& Representation of the $n$th coming instance\\
         $Y_n$& True Label of the $n$th coming instance\\
         $Y'_n$& Predicted label of the $n$th sample of coming instances\\
         $C_t$& Class of the $t$th leaf model\\
         $B$& Size of Collection set $\mathcal{B}$\\
         $H$ & Length of the extracted feature vector of each instance\\
         $F$ & Size of reference set\\
         $\mathcal{K}_t$ & Number of known classes at Stage t\\
         $\mathcal{B}$ & Collection set of newly detected instances\\
         $\mathcal{F}$& Reference set with fixed size\\
         $\theta$ & Threshold for selecting rejection line for leaf node\\
         \hline\hline
    \end{tabular}
    \label{tab:my_label}
\end{table}

As new classes appear continuously and must be detected and incorporated accordingly, we define an OWR process consisting of several stages, with a new class detected and learned at each stage. At a Stage $t$, let the classes that are included in the recognizer as known classes be labeled by the positive integers $\mathcal{K}_t=\{1,2,...,C_t\}\subseteq \mathbb{N}^{+}$. Instances from either known or unknown classes are represented as $(X_n,Y_n)$, where $X_n$ is the $n$th instance of all the incoming instances and could be represented as either a matrix (e.g., an image) or a vector (e.g., a time series of objects). $Y_n \in \mathcal{K}_t$ if $X_n$ is from known classes, and $Y_n=0$ if $X_n$ is from unknown classes. In this scenario, some instances come with an identity function $I$ which provides side-information indicating whether two instances are from the same class or not, i.e., $I(X_n,X_m)=0~or~1$, where 1 indicates that instance $X_n$ and $X_m$ belong to the same class and 0 that they do not. The open world recognition problem contains three tasks: classifying the known classes, detecting newly emerging classes and learning the new classes. Therefore, the problem is defined as learning a dynamic recognition function $M_t:X_n \rightarrow Y_n\subseteq \mathbb{N},\ t=1,\cdots$ which contains unique feature patterns that are used to recognize known classes and detect unknown classes. Furthermore, the recognition function $M_t$ should be scalable to recognize all the classes that have ever been seen so that at each stage it is possible to scalably update the recognition mapping from Stage $t$ to Stage $t+1$ as: $M_t \rightarrow M_{t+1}$ using the dynamic features to include the new class.

However, to handle all the solutions to the end-to-end OWR problem, several challenges must be considered: 1)~It is difficult when performing an end-to-end feature extraction in a fixed architecture to both include the features for a newly detected class and maintain the previously learned features for distinguishing known classes, especially considering the memory and time consumption. 2)~It is difficult to ensure the updated features do indeed distinguish the newly added class from the potential future unknown classes. 3)~It is difficult to utilize the relationships among the instances indicated by $I$ to assist the feature extraction and maintain the consistency of feature patterns for instances with the same class. 

\section{The proposed MDCC}
In this section, we propose the Multi-stage Deep Classifier Cascades (MDCC) to address the above challenges in the open world recognition scenario. First, an introduction of the overall architecture and the incremental process of the cascades are given. Then, a more detailed consideration of the proposed deep one-class classifier and a self-describing regularization mechanism are presented.

\begin{figure*}[htb]
\centering
\includegraphics[width=0.7\textwidth]{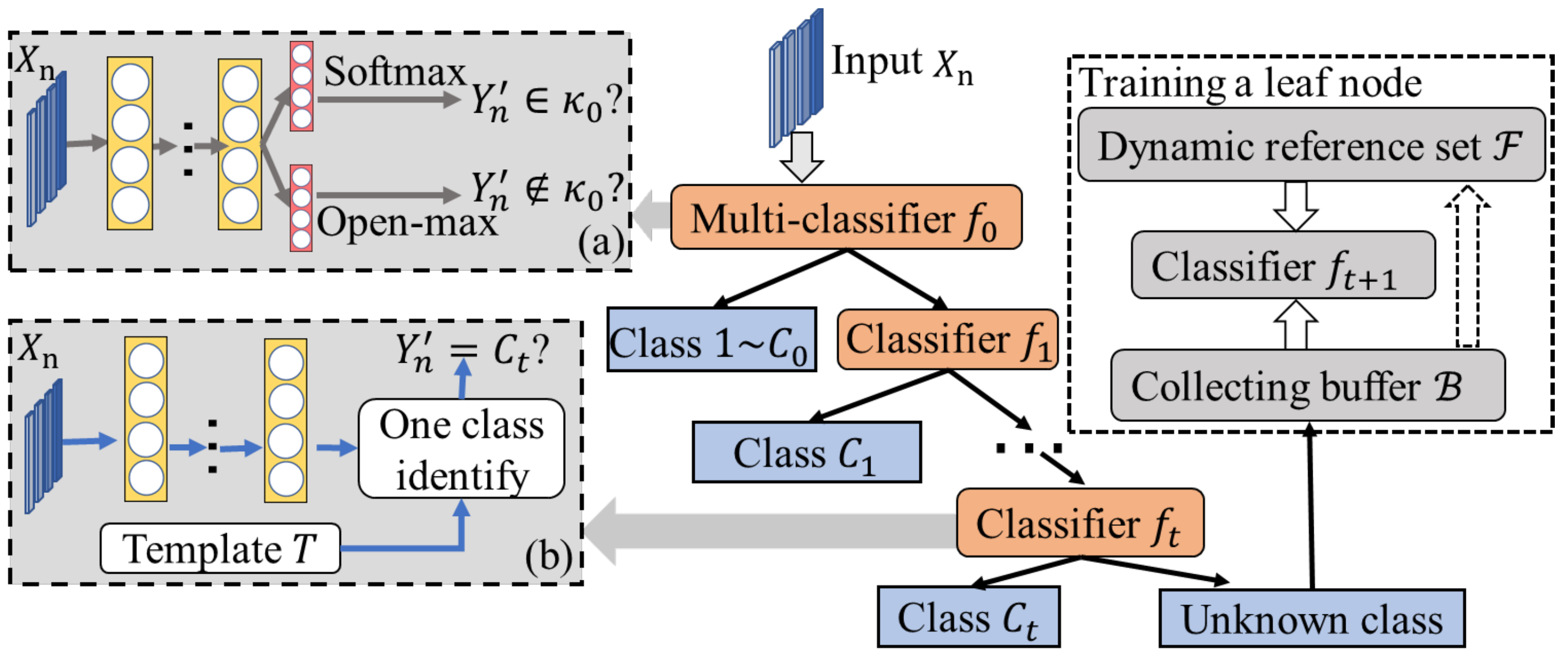}
\caption{Architecture of Multi-stage Deep Classifier Cascades and model details of (a) root node and (b) leaf nodes}
\label{fig:architecture}
\end{figure*}

\subsection{Overall Architecture}
\textbf{Multi-stage cascade architecture}. 
To incrementally detect and learn new classes for the open world problem, we propose a novel Deep Convolution Neural Network (DCNN) based cascade architecture which consists of a root and several leaf classifiers trained in an end-to-end fashion. It can either reject the instance as a new class or classify the accepted instance as a member of the known classes. The DCNN is chosen as the base model due to its excellent performance and general ability to deal with input instances in any form. The merits of the overall recognition model are as follows: 1) It can contain sufficient and unique features for distinguishing among all the known classes at any stage. 2) It can increment the leaf nodes to both recognize a newly added class as well as detect future unknown classes. 3) It can learn and include new features efficiently without disturbing the existing features.

Fig.~\ref{fig:architecture} shows the overall architecture of the proposed MDCC. At Stage $0$, the function $M_0$ is realized by a root node $f_0$ to classify the existing known classes and detect unknown classes, as shown in the orange rectangle in Fig. \ref{fig:architecture}. At Stage $t$, once an unknown class is detected by the function $M_t=[f_0,...,f_t]$ and sufficient instances for this new class are collected in a buffer, a one class classifier $f_{t+1}$ is trained to learn the feature information of the new class and it is added as a leaf node of the cascade. The recognition function is then updated as $M_{t+1}=[f_0,...,f_{t+1}]$. The loss minimized at Stage $t$ is defined as follows:
\begin{align}
\mathcal{L}(M_t)=
\left\{
\begin{array}{lll}
\mathcal{L}_{root}(f_0) && t=0\\
\mathcal{L}_{leaf}(f_t) && t>0,
\end{array}
\right.
\end{align}
where $\mathcal{L}_{root}(f_0)$ is the cross entropy loss for training the root classifier, and $\mathcal{L}_{leaf}(f_t)$ is the overall loss function of the leaf classifier $f_t$, which is described in more detail in Section.\ref{section:one class}. Optimization methods (e.g., Stochastic gradient descent and the Adam algorithm) based on Back-propagation technique can be utilized to optimize each node model.

To recognize an instance $X_n$ at Stage $t$ via the function $M_t=[f_0,...,f_t]$, $X_n$ is input into each of the node models in turn until it is accepted by a node. The input of each leaf node is the instance rejected  by the previous node. Specifically, at root node $f_0$, if $X_n$ is accepted as known, it will be classified by $f_0$; Otherwise, it enters the next node. At a leaf node $f_i (i<t)$, if $X_n$ is accepted, it is predicted as Class $C_i$; Otherwise, it enters the next node. If $X_n$ is finally rejected by the last node $f_t$, then it is identified as being of unknown class at stage $t$ and is stored in a buffer. All the instances detected as new classes will be assigned into the various buffers based on their identity function. Thus, each buffer $\mathcal{B}$ contains the instances of the same class with which to train a leaf node. The detailed process for the open world recognition of instance $X$ at Stage $t$ are presented in Algorithm.~\ref{alg}.

\begin{algorithm}[htb]
\caption{Single instance recognition process for open world recognition at Stage $t$.}
\label{alg}
\begin{flushleft}
\textbf{Input}: The incoming instance $X_n$\\
\textbf{Requires}: Current cascade model $M_t=[f_0,...,f_t]$.\\
\textbf{Output}: The predicted label $Y'_n$ of $X_n$.
\end{flushleft}
\begin{algorithmic}[1]
\STATE Let $i=0$.
\WHILE{$i<=t$}
\STATE $Y'_n=f_i(X_n)$ .
\IF {$Y'_n\in \mathcal{K}_t$}
\STATE \textbf{Return} $Y'_n$.
\ELSE
\STATE i=i+1
\STATE Continue.
\ENDIF
\ENDWHILE
\IF {i>t}
\STATE{\textbf{Return} $Y'_n=0$}
\STATE{Train leaf node $f_{t+1}$}
\STATE{Update $M_t$ to $M_{t+1}=[f_0,...,f_{t+1}]$}
\ENDIF
\end{algorithmic}
\end{algorithm}

\textbf{Root node at initial stage}.
To extract features and classify the multiple known classes at the initial stage, a DCNN is utilized for the multi-class classification as the root node. This DCNN consists of several convolution layers, average pooling layers, and fully connected layers, as well as the SoftMax layer used to recognize the known classes. In addition, to detect any new classes, the openmax layer proposed by~\cite{bendale2016towards} is utilized in parallel with the softmax layer, as shown in Fig.\ref{fig:architecture} (a). This openmax layer helps to decide whether to accept the instance as a known class or not and, once accepted, the instance is then classified by the softmax layer. Specifically, the softmax layer outputs the probability of the instance belonging to each known class $P(Y'_n=k|X_n,k \in \mathcal{K}_0)$. While the openmax layer outputs the probability of $X_n$ being the unknown class as $P(Y'_n=0|X_n)$.

The output of the openmax layer is computed based on the distances between the output of the activation layer (the last layer before the SoftMax layer) and $\mu_{k}(k\in \mathcal{K}_0)$ for each class $k$ through a Weibull model. First, $\mu_{k}$ is computed during the training phrase as the mean of the activation outputs of all the training instances belonging to class $k$. Then a Weibull
model obtained through libMR~\cite{chang2011libsvm} is used to model the distribution of $\mu_{k}$ and the largest distance between
$\mu_{k}$ and all the correctly classified samples. Based on the Weibull model generalized, we can compute the Weibull CDF probability on the distances between $P(Y'_n=0|X_n)$ and $\mu_{k}$.
Since the probability is expected to be meaningful only for a few top ranks, we compute the weights for the top ranking classes and use these to scale the Weibull CDF probability. The
$P(Y'_n=0|X_n)$ will be revised based on the changed scores. For the new class, a pseudo-activation layer is computed to keep the total $P(Y'_n=0|X_n)$ value constant. The details of the Weibull model and openmax computation can be found in~\cite{scheirer2011meta,bendale2016towards}. 

\textbf{Leaf nodes with dynamic reference set}. We propose to train a deep one-class classifier that both recognizes the newly detected class and detects the future new classes. As each new class is detected, it is necessary to update the current recognizer to include this new class without retraining the model. The one-class classifiers only need to learn the features of the new class and can reject all the instances not belonging to this class. 
To learn the features that not only describe the new class but also distinguish it from future unknown classes, each one-class classifier incorporates an end-to-end feature extraction part and a distance-based classifier. 
This one-class classifier is trained based on a proposed dynamic reference set in order to both maintain the performance and minimize memory consumption.

\textbf{Self-describing regularization}.
In the one-class classifier training, to further minimize the feature space required to largely distinguish the detected new class from others and fully utilize any side-information on the instances' relationships, we propose a self-describing regularization by enforcing the feature consistency of collected instances in the buffer. The instances are collected based on the side-information they contain regarding the identity function indicating the relationships among the instances of unknown classes.


\subsection{Deep one-class classifiers as leaf nodes}
\label{section:one class}
To learn the features required to describe the newly detected class, we propose the one-class classifiers as leaf nodes trained in an end-to-end fashion, as shown in Fig.~\ref{fig:architecture} (b). The leaf nodes in the MDCC are incrementally added at each stage to learn and include the new class' information, simultaneously detecting any future unknown classes simultaneously. The one-class classifier consists of two parts: a feature extractor and a distance-based classifier. For convenience, here the newly detected class is named the \emph{target class} for training a classifier.

\textbf{End-to-end Feature extraction}. The feature extraction part of a leaf node $f_t$ is based on a DCNN model $g_t$ at Stage $t$, which is trained by improving the outer-describing property of the feature space. The outer-describing property refers to the ability of features to fully describe various classes. Improving the model's outer-describing ability can minimize the feature space needed to describe the target class and, thus, improve the distinguishability of the target class from other unknown classes. Specifically, a reference set $\mathcal{F}=\{(X^{(\mathcal{F})}_n,Y^{(\mathcal{F})}_n)\}_{n=1}^F$ consisting of instances from various existing known classes is utilized to enlarge the outer-describing ability for describing the newly detected class at each stage. $F$ is the number of instances in $\mathcal{F}$. The reference set first collects instances of each known class at Stage $0$ and then continues to included instances of more classes at each stage. The outer-describing loss for training $g_t$ can thus be minimized as follows:
    \begin{equation}
        \mathcal{L}_{outer}=\mathbb{E}_{X^{(\mathcal{F})}_n,Y^{(\mathcal{F})}_n}[-\sum\nolimits_{n=1}^{F}log(g_t(X^{(\mathcal{F})}_n))].
    \end{equation}
During the testing phase, the features for one instance can be extracted from the layer before the softmax.

\textbf{Error-corrective distance based classifier}. An error-corrective distance based classifier is proposed for the extracted features to evaluate whether an instance is from the target class or not. Since the features extracted by $g_t$ only have the ability to describe various classes, its softmax layer cannot distinguish the new class from any future unknown classes. To deal with this, each instance is evaluated based on the distance $D(h_t(X_n),\upsilon_t)$ between its features $h_t(X_n)$ and the mean feature vector $\upsilon_t$ of the training instances in the buffer $\mathcal{B}=\{(X^{(\mathcal{B})}_n,Y^{(\mathcal{B})}_n)\}^{B}_{n=1}$, where $B$ refers to the number of instances. $h_t(X_n)$ is the feature vector outputted from the feature extraction layers (before the softmax layer) in $g_t$. $\upsilon_t$ is computed as $\sum\nolimits^B_{n=1}D(h_t(X^{(\mathcal{B})}_n),\upsilon_t)/B$. The distance function $D$ can be any distance measurement, for example the cosine distance. Then the maximum distance $d_t=max \{D(h_t(X^{(\mathcal{B})}_n),\upsilon_t)\}_{n=1}^{B}$ based on the training instances in buffer $\mathcal{B}$ is used as the rejection line. In addition, to minimize the influences of errors generated from the last leaf node's recognition, we allow $\theta$ percent of instances to be outlier instances and filter out the largest $\theta$ percent of distances, selecting the rejection line as $d^{(\theta)}_t$. Thus, the predicted label for an object $X_n$ in the leaf node $f_t$ is computed as follows:
\begin{align}\nonumber
\label{equ3}
f_t(X_n) =
\begin{cases}
unknown & \text{if}\quad D(h_t(X_n,\upsilon_t)>=d^{(\theta)}_t\\
C_{t} & \text{if}\quad D(h_t(X_n,\upsilon_t)< d^{(\theta)}_t.
\end{cases}
\end{align}

\textbf{Dynamic Reference Sets}. As new classes arrive continuously, to maintain the diversity of the outer-describing property and limit memory consumption, a dynamic reference set with a fixed size rather than one that is incrementally enlarged is proposed. On one hand, the more classes that are contained in the reference set for training, the better outer-describing ability the feature space will have, so after a new leaf node is built, the instances from the newly added class need to be included in the dynamic reference set. However, on the other hand, to avoid a rapid escalation in the computation and memory costs due to the increasing size of the reference set, we must restrict the total number of instances in the set by reducing the number of instances of each class. The number of instances for each class in the reference set is computed as: $F/C_t$, where $F$ is the fixed size of the reference set. The reduced instances for each class are randomly chosen.

\subsection{Self-describing regularization}
Feature space regularization is proposed due to two considerations: 1) to improve the ability of the extracted features to describe the target class, which is referred to as its self-describing property; and 2) to remedy the deteriorating performance of end-to-end feature extraction due to the decresing number of of instances of each class in the dynamic reference set. As instances of new classes are mixed, instances from the same new class must be minimized and instances from different classes are maximized. Thus, for any two instances $X_n$ and $X_m$ detected as unknown classes, the distance between their features should be minimized as: $min~I(X_n,X_m)||h_t(X_n)-h_t(X_m)||_2$, where $I$ is the pre-knowledge indication of the relationships between two instances. Since several instances with the same class are collected in $\mathcal{B}$ with the property of $I(X^{(\mathcal{B})}_n,X^{(\mathcal{B})}_m)=1$, the self-describing regularization for training a one-classifier is minimized, which is expressed as follows:
\begin{equation}\nonumber
    \mathcal{R}(X_n)=\frac{1}{BH} \sum\nolimits_{n=1}^{B} (h_t(X^{(\mathcal{B})}_n)-\omega_t)^{T}(h_t(X^{(\mathcal{B})}_n)-\omega_t),
\end{equation}
where $H$ is the length of the extracted feature and $\omega_t$ is the mean feature vector of all the instances' features in $\mathcal{B}$. Then the overall loss function of a leaf node $f_t$ is computed as:
\begin{equation}\nonumber
\begin{split}
   \mathcal{L}_{leaf}(f_t)=\mathbb{E}_{X^{(\mathcal{F})}_n,Y^{(\mathcal{F})}_n}[-\sum\nolimits_{n=1}^{F}log(h_t(X^{(\mathcal{F})}_n))]\\
   +\beta\frac{1}{BH} \sum\nolimits_{n=1}^{B} (h_t(X^{(\mathcal{B})}_n)-\omega_t)^{T}(h_t(X^{(\mathcal{B})}_n)-\omega_t),
   \end{split}
\end{equation}
where $\beta$ controls the trade-off between the $\mathcal{L}_{outer}$ and $\mathcal{R}(X_n)$. For optimization, the gradients minimizing the two components of the loss function are computed separately.

It is worth noting that the side-information requirement is not a limitation of the proposed MDCC. First, the existing works typically ignore this assumption and implicitly assume the new classes are unmixed or the samples arrives in a batch. Instead, our method utilizes this prior knowledge by proposing a self-describing regularization and yields better experimental results, as shown in the next section. Second, from another point of view, the proposed MDCC could be categorized as multi-instance learning based on knowing the relationships among the samples. Thus, MDCC is a general method and can include the case of single-instance learning without the need for an indicating function.



\section{Experiment}
In this section, the effectiveness of the proposed MDCC is evaluated for two real-world datasets and the results obtained compared with those of three existing state-of-the-art method in the domain of open-world recognition. The parameter sensitivity of threshold $\theta$ used in the one-class leaf model was analyzed to validate the robustness of the MDCC. All the experiments were conducted on a 64-bit machine with Intel(R) Core(TM) quad-core processor (i7CPU@ 3.40GHz) and the NVIDIA GPU GTX1070. 

\begin{figure}[h!b]
\centering
\includegraphics[width=.45\linewidth]{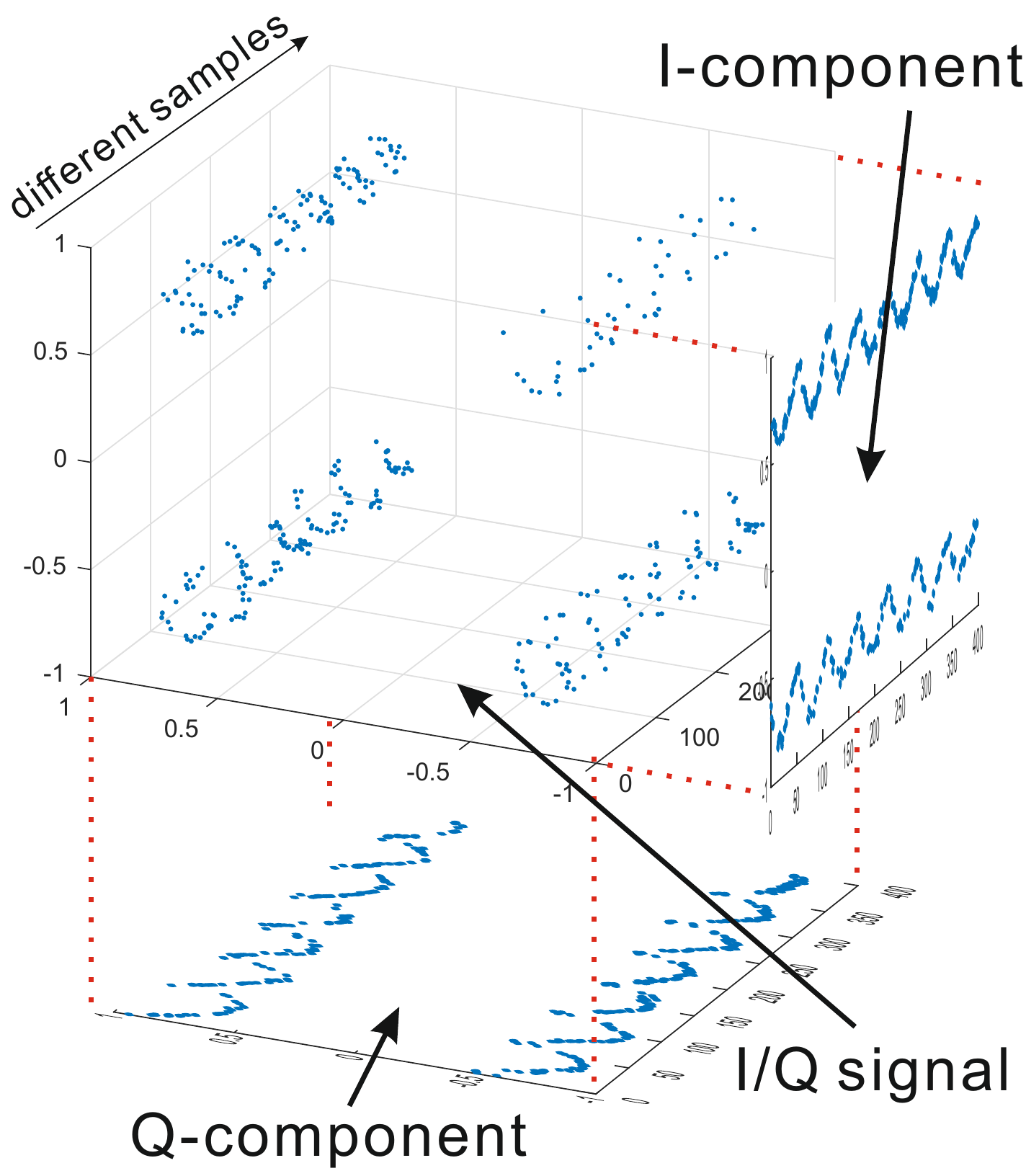}\vspace{-0.2cm}
\caption{The I/Q data is illustrated as a 3-D graph; the 2-D graphs are two projections of the 3-D graph. When I/Q data are considered as two separate series of real numbers, the underlying dependency between them will fail to be utilized.}
\label{fig3}
\end{figure} 

\subsection{Experimental Setup}
\subsubsection{Datasets} Two real-world datasets were used to validate the performance of the proposed MDCC\footnote{The data and the code are available at: https://github.com/xguo7/MDCC-for-open-world-recognition}. The process of data collection and the experiment settings utilized are detailed in this section.

\textbf{RF signal Datasets}. We collected RF signals to test the effectiveness of our method for real world applications. The objective of RF signal recognition is to recognize different RF devices based on
the signals they transmitted. Each sample bag is
a time-series signal which has complex values consisting of a real in-phase component (I) and an imaginary quadrature component (Q), as shown in Fig. 3. Each signal sample also comes with an identity label representing the unique device (e.g., transmitter) it originates from. For RF signal collection, we set up a wireless line-of-sight communication system with a static transmitter and receiver. The distance between the receiver and transmitter was fixed at 50cm. The same USRP-2943R device was utilized as the receiver and fix it through-out the entire testing process. In the receiver part,
The receiver recieves the transmitted modulated signal and down converts it through the RF circuits before sending it on to the LabVIEW software for recording. We used two USRP-2943R and two USRP-N210 devices as transmitters in this experiment. This dataset was chosen originally due to the practical reasons and the difficulty of feature extraction tasks.

Eight sets of RF signal samples were collected and labeled from 1 to 8 in turn. The signal series for each class consisted of 4,096,000 data points for each of 4 columns, representing the values for the real and imaginary parts of both the received IQ sample and the associated standard. The data was split into 5,000 samples, with each sample consisting of 512 data points. Among the 5,000 samples of each class, 1,000 were collected in the buffer as the training set, with the indication that they are from the same long signal series, and the remainder used for testing.

\textbf{Twitter dataset}. The proposed MDCC was also validated for s Twitter topic recognition task. Here the data were collected using Twitter's Streaming API (filter/track method) to obtain a stream of public tweets filtered according to a given keyword set. Each collected tweet contains metadata along with the tweet message content, including the time of posting, location of source if available, and author profile information such as the author profile description, and number of followers and friends, number of statuses, across 11 days from 27 October 2012 to November 7 2012. Due to the limitations of Twitter's API policies, we utilized 4.9 million tweets borrowed from a labeled sample of tweets originally used in a prior study published in the crisis informatics literature~\cite{purohit2014emergency}, where the labeled classes included the following: {Clothing, Food, Medical supplies including blood, Money, Shelter, Volunteer work}. The labels were obtained using a crowd sourcing approach. Each collected tweet was then represented by an average word embedding vector using pretrained Twitter-GloVe embeddings~\cite{pennington2014glove} as inputs, each with a length of 200. 80 samples of each class were randomly chosen and assigned an identification to be collected in a buffer as training samples, while a further 80 samples of each class were used for testing.

\subsubsection{Evaluation protocol}
In the open world problem, the training
phase is a complex process as new classes are continually being added to update the model. The open set evaluation protocol proposed by~\cite{bendale2015towards} needs certain unknown labels to validate several parameters and is thus not suitable for open-world problem. An online learning protocol by~\cite{mu2017classification} does not consider the error propagation. We therefore propose a new protocol that can reflects real world scenario s more realistically.

\textbf{Training Phase:} For the training phase, training samples for each class were mixed together. Samples from two classes were deemed known for training the root node, with samples from new classes arriving one by one to be detected and learned. For example, in Stage 1, classes 1 and 2 are known, and class 3 is unknown. And we only test the stages where both known classes and unknown classes exist. The samples detected as new and placed in class 3 are then used to update the model. 

\textbf{Testing Phase:} There are additional testing samples included for each class. The testing is conducted at every stage after the model has been updated. All the testing samples are used for every test. For example, in Stage 3 all the trained child models (initial model, one class model 1 and one class model 2) are tested by the combination of known samples and unseen samples.

To evaluate the performance, we utilize two measures, EN-Accuracy and F-measure~\cite{mu2017classification}. EN-Accuracy is computed as:
\begin{align}
 EN-Accuracy=\frac{(N-known+N-unknown)}{N},   
\end{align}
where N is the total number of testing samples. N-known and N-unknown are the number of correctly recognized samples with known classes and with unknown classes, respectively. 
The F-score is defined as: 
\begin{align}
  F-score=\frac{2TP}{(2TP+FP+FN)},
\end{align}
where TP (True Positive) represents the correctly classified samples of unknown classes, FP (False Positive) is the number of incorrectly classified samples of unknown classes, and FN (False Negative) represents the number of incorrectly classified samples for the known classes.

\subsubsection{Comparison algorithms}
To validate the priority of the
proposed method, several classical methods were also included here for comparison. A brief description of the methods is as follows:
\begin{itemize}
    \item 1) Local novel detector (LOD) \cite{bodesheim2015local} is a multi-class novel detection method that can be combined with multi-class SVM~\cite{chang2011libsvm} classifier to solve the open world problem.
    \item 2) R-Openmax: Openmax~\cite{bendale2016towards} was combined with a retraining policy for incremental learning. \item 3) S-Forest~\cite{mu2017classification} is an online learning method ofen used to deal with streaming data. During the incremental process, it collects new samples and updates the model once the number of samples reaches a certain threshold.
\end{itemize}

For methods like LOD and S-Forest, features of RF signals have to be manually extracted. 12 classical indicators of signals were therefore used as the features, namely the Mean, Median, Median Absolute Deviation, Standard Deviation, Skewness, Kurtosis, Max, Min, Mean Square, Root Mean Square, Pearson-Skewness, and Mean Absolute Deviation.

\subsubsection{Architecture of the base DCNN model.}
\begin{figure}[htb]
    \centering
    \includegraphics[width=0.9\linewidth]{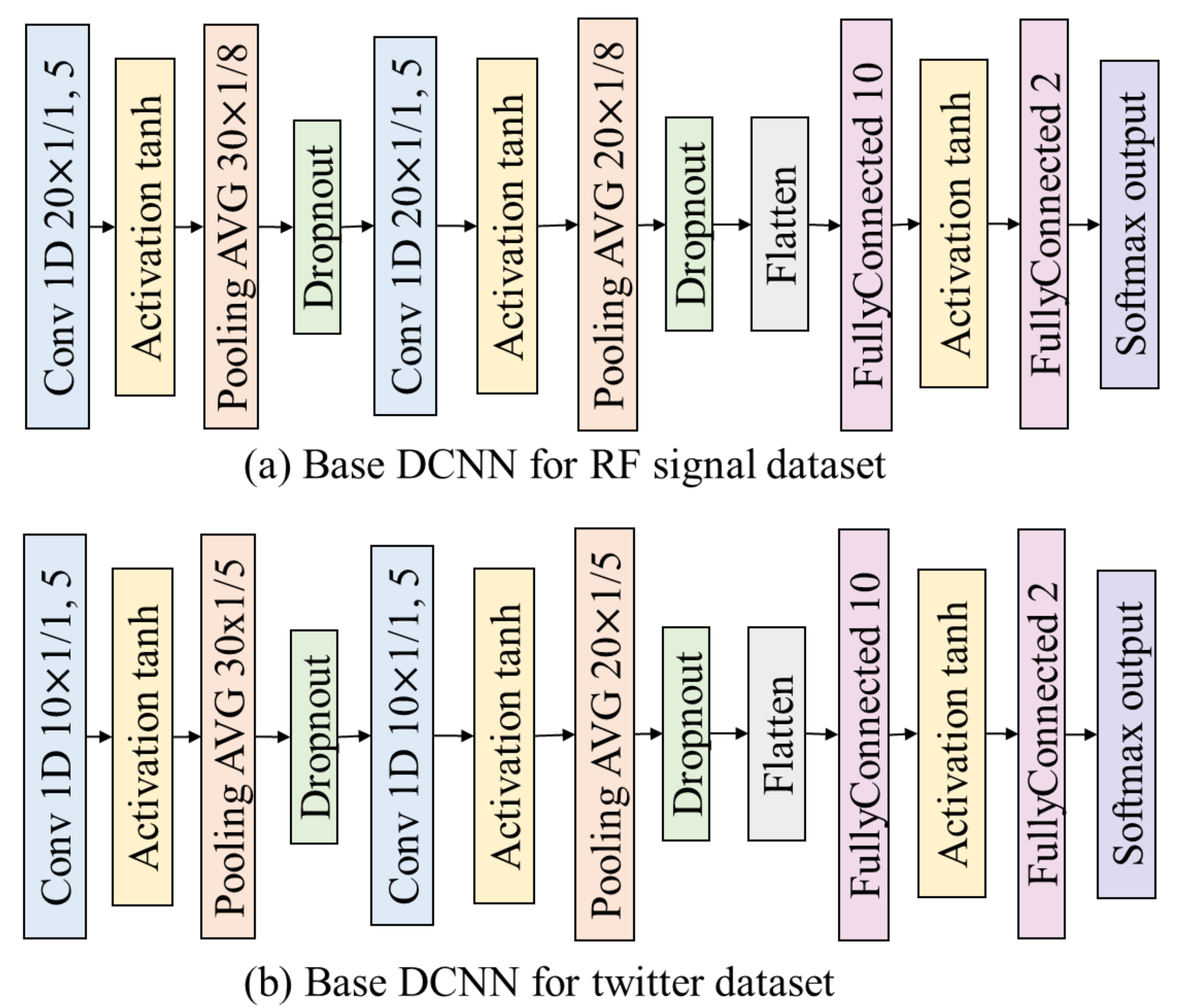}
    \caption{Architecture details of DCNN base model used in the two datasets.}
    \label{fig:baseadcnn}
\end{figure}

Fig.~\ref{fig:baseadcnn} shows the architecture of the base DCNN model used as the root and leaf node in the proposed MDCC for the two real-world datasets. Both root and leaf nodes share the same architecture in the MDCC. The openmax layer of the root node is eliminated here since it is computed based on the output of the softmax layer. The structure of the convolution layer is expressed as: $<\!filter~size/strides,~ number~of~filters\!>$. The structure of the pooling layer is expressed as: $<~\!pooling~size/strides\!>$. The dropout probability of each layer is 50\%.

\subsubsection{Parameters used in the experiments} Table~\ref{table:notations} lists the parameters used in the training and testing phase in the experiments. $lr$-root and $lr$-leaf refer to the learning rate of optimization for the training root model and leaf model, respectively. $\alpha_{root}$ refers to the number of “top” classes that are chosen for revision in order to compute the openmax output in the root model. $\gamma_{root}$ refers to the rejection threshold of the openmax layer to define the new class of root model. The details of these two parameters can be found in~\cite{bendale2016towards}. The term $\theta$ refers to the estimated percentage of outlier instances when building the classifier for the leaf node, which determines the rejection line of the new classes.

\begin{table}[htb]
  \centering
  \caption{Parameters used in the experiments\vspace{-0.4cm}}
  \begin{tabular}{lllllll}\\
  \hline\hline
    Dataset& $lr$-root& $lr$-leaf & batch-size & $\alpha_{root}$ &$\gamma_{root} $ &$\theta$\\
    \hline
     RF signal&0.005 &0.002 &50 & 2&0.008&0.7  \\
      \hline
     Twitter&0.001 &0.002 &20 &2 &0.008&0.5 \\
    \hline\hline
  \end{tabular}
  \label{table:notations}
\end{table}

\subsection{Experiment Performance}
\subsubsection{Convergence of loss functions}
\begin{figure*}[htb]
\centering
\includegraphics[width=\linewidth]{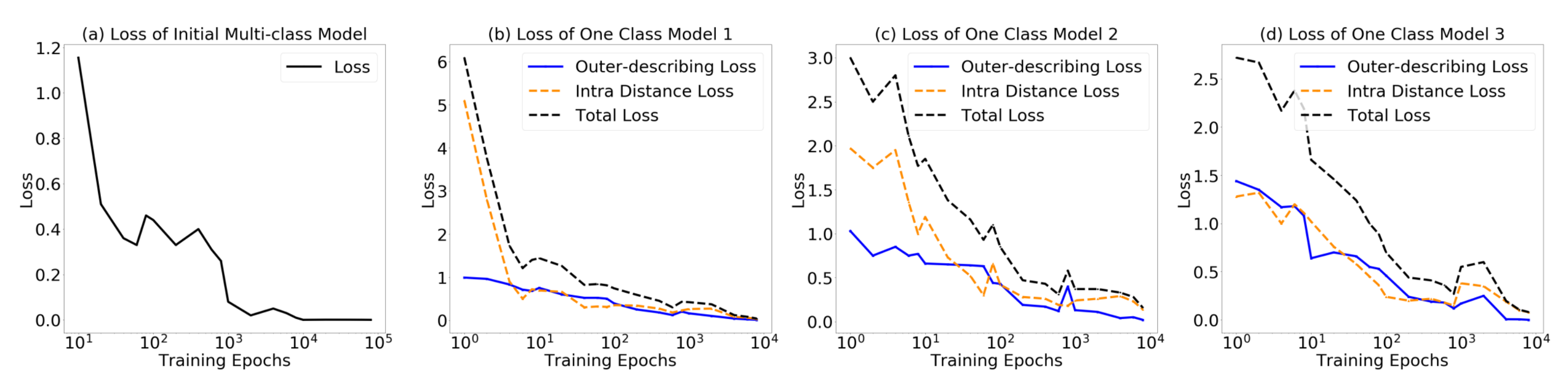}
\caption{Convergence of loss functions}
\label{fig:Convergence of loss functions}
\end{figure*}

There are two kinds of loss incurred in training a MDCC model: outer-describing loss and intra distance loss (i.e.self-describing regularization). To show the convergence of the two kinds of loss and validate the effectiveness of the training strategy, we recorded the convergence process for the outer-describing loss, intra-distance loss and their total loss. Figure. \ref{fig:Convergence of loss functions} shows the loss convergence process for the root model and three newly added leaf models. The outer describing loss and intra-distance loss convergence synchronously and experience a steady decrease to around 0, after around 10,000 iterations. These results confirm that the MDCC with a self-describing regularization can be well trained to minimize both the outer-describing loss and intra-distance loss.

\subsubsection{Performance evaluation on RF signal dataset}
Table \ref{table:comparsion result on signal} shows the EN-Accuracy and F-score evaluated from Stage 1 to 6 by the proposed MDCC and the comparison methods. These results validate the effectiveness of MDCC, which achieved the best performance in 75\% of the stages for both the EN-Accuracy and F-score; it also exhibited a stable performance as the number of classes increased. Specifically, MDCC outperformed both LOD and S-Forest by 29.1\% and 23.2\%, respectively, in EN-Accuracy, and 19\% and 37.6\%, respectively, in F-score. This because both S-Forest and LOD utilize handcrafted features defined initially, while MDCC updates and extracts dynamic features through supervised learning by a deep model. MDCC achived a performance that was similar to that of R-openmax during the first stage, but then went on to outperforms it by 13.4\% in EN-Accuracy and 49.5\% in F-score. This is because although R-openmax has a high feature extraction ability, the model tends to deteriorate quickly as new classes emerge since its current features are insufficient to learn more new classes. In addition, these results show that the cascade framework of the proposed MDCC contributes greatly to its overall performance.

\begin{table}[htb]
  \centering
  \caption{EN-Accuracy (EN-Acc) and F-score from Stage 1 to Stage 6 for the different methods tested on the RF signal dataset\vspace{-0.5cm}}
  \begin{tabular}{llrrrr}\\\hline\hline
    Metric &Stage &LOD & S-Forest &R-openmax &MDCC \\\hline
    \multirow{6}*{EN-Acc(\%)}& 1 &36.95 &45.77 &\textbf{60.45} &\textbf{60.45}\\
    ~& 2 &34.85 &33.24 &43.08 &\textbf{57.32}\\
    ~&3 &53.35&37.03&42.75 &\textbf{54.50}\\
    ~& 4 & 33.83&55.20 &\textbf{58.66}&50.43\\
    ~& 5 &25.83& 33.68&54.39 &\textbf{55.65}\\
    ~& 6 &31.87& 32.38&33.55  &\textbf{36.55}\\
    \hline
    \multirow{6}*{F-score}& 1 &\textbf{0.50}& 0.33&0.41 &0.41\\
    ~& 2 & 0.50&0.27& 0.29&\textbf{0.75}\\
    ~& 3 & 0.60&0.19& 0.17&\textbf{0.62}\\
    ~&4&0.34&\textbf{0.62}&0.31&0.42\\
     ~&5&0.18&0.11&0.27&\textbf{0.40}\\
      ~&6&0.14&0.11&0.10&\textbf{0.19}\\\hline\hline
  \end{tabular}
  \label{table:comparsion result on signal}
\end{table}

\subsubsection{Performance evaluation on the Twitter dataset}
Table~\ref{table:comparsion result on twitter} shows the performances in terms of EN-Accuracy and F-score from Stage 1 to 4 for the Twitter dataset achieved by the different methods. In general, MDCC achieved the best performance in 75\% of the stages in both  EN-Accuracy and F-score. Specifically, MDCC outperformed LOD and S-Forest by 18.5\% and 6.25\%, respevtively, for the EN-Accuracy, and 32\% and 39.7\%, respectively, for the F-score. This is because MDCC can update and extract the features through end-to-end supervised learning by deep neural networks for this difficult recognition task and takes into account the errors generated from the last stage. MDCC performed equally to R-openmax in the first stage, but then subsequently outperformed it by 13.4\% in EN-Accuracy and 39.5\% in F-score. This is because the predefined architecture of R-openmax limits the types and number of features used to distinguish incoming new classes.

\begin{table}[htb]
  \centering
  \caption{EN-Accuracy (EN-Acc) and F-score from Stage 1 to Stage 4 for the different methods tested on the Twitter dataset\vspace{-0.5cm}}
  \begin{tabular}{llrrrr}\\\hline\hline
    Metric &Stage &LOD &S-Forest &R-openmax &MDCC \\\hline
    \multirow{4}*{EN-Acc(\%)}& 1&37.50 &43.12 &\textbf{49.53} &\textbf{49.53} \\
    ~& 2 &18.33 &35.00 &40.70 & \textbf{44.78}\\
    ~&3 &42.01&33.44&\textbf{42.02} &35.77\\
    ~& 4&31.50 & 34.27&34.50&\textbf{35.90}\\
    \hline
    \multirow{4}*{F-score}& 1& 0.27 &0.31 & \textbf{0.78}&\textbf{0.78}\\
    ~& 2 & 0.31&0.28&0.61 &\textbf{0.62}\\
    ~& 3 & 0.39&0.25& \textbf{0.40}&0.37\\
    ~&4&0.16&0.19&0.15&\textbf{0.22}\\\hline\hline
  \end{tabular}
  \label{table:comparsion result on twitter}
\end{table}

\subsubsection{Sensitivity of the Classifier Threshold}
\begin{figure}[htb]
\centering
\includegraphics[width=\linewidth]{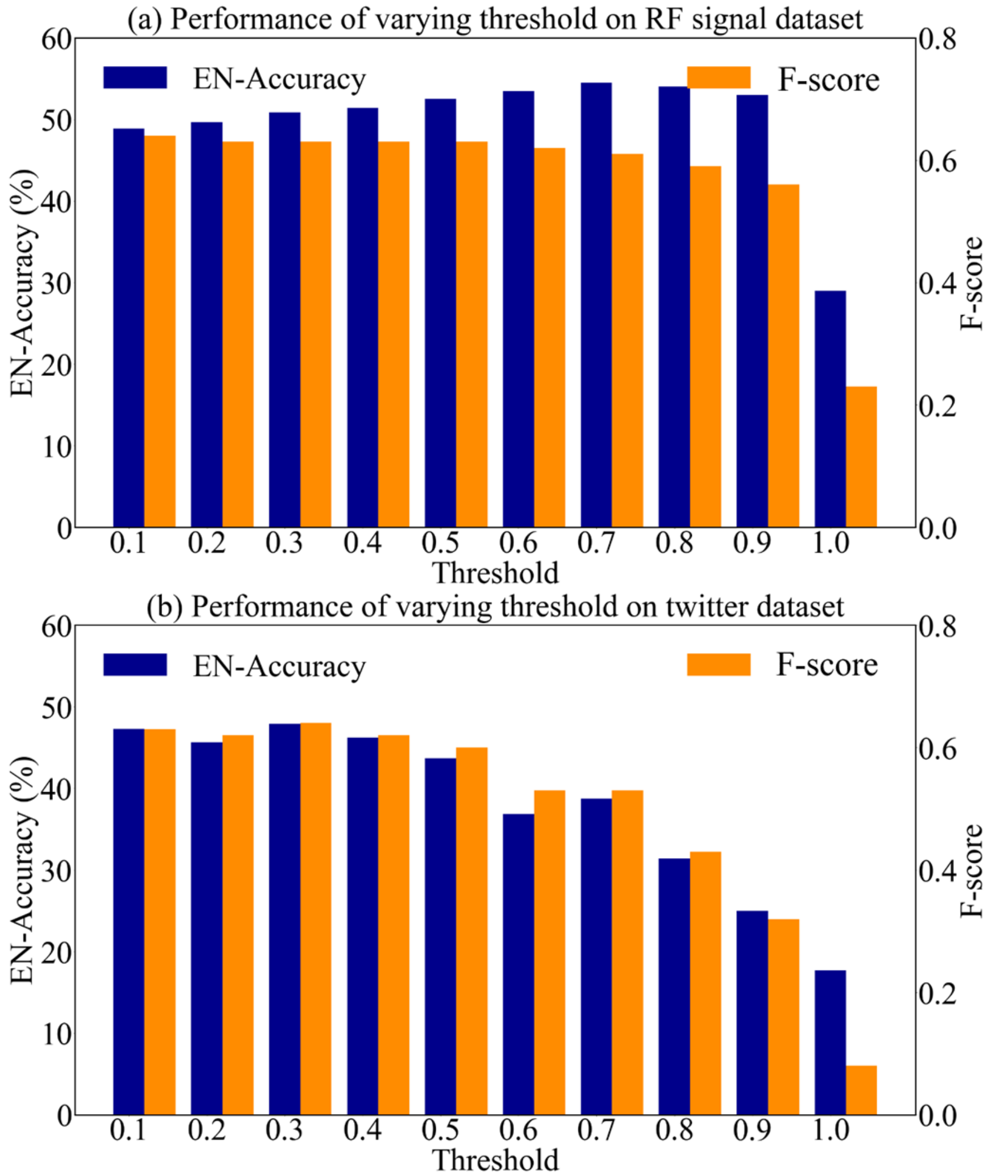}
\caption{EN-Accuracy and F-score for different thresholds for (a)~RF signal dataset and (b)~Twitter dataset.}
\label{fig5}
\end{figure}

There is one main parameter in the proposed MDCC, namely the percentage threshold $\theta$ in the leaf node used for filtering the outlier instances, which determines the rejection line of new classes. We used the same initial conditions and datasets as those mentioned before to test the parameter sensitivity. Fig.~\ref{fig5} shows the sensitivity results in Stage 3 of varying $\theta$ in two datasets. Stage 3 was chosen because the number of known and unknown classes at that stage are balanced. The threshold $\theta $ varies from 0.1 to 1. In general, for both datasets, the F-score and EN-Accuracy are slightly lower than for the other when $\theta>0.5$, but when it is between 0.1 and 0.5, the result becomes stable. This is because the presumption that the outlier samples will be generated from the last node is over-estimated and more than half of the detected new samples are filtered. Overall, the threshold $\theta$ remains stable and thus achieves a good performance over a reasonable range between $0.1\sim 0.5$.

\section{Conclusions}
In this work, we addressed three challenges by proposing the end-to-end Multi-stage Deep Classifier Cascades for open world problem. First, the multi-class cascade architecture was implemented to incrementally detect and learn new coming classes. Second, an error-corrective one class classifier trained based on a dynamic reference set is proposed, where the leaf nodes learn unique features for each new coming class. Third, a feature space regularization based on the collective target set was utilzed to enforce the feature consistency of all instances in the same target set. Two real world experiments on the comparison to several existing methods indicate the effectiveness and efficiency of the proposed method. Tests to analyze the model parameter threshold indicated that a stable performance was achieved when the threshold was within a reasonable range.

\section*{Acknowledgement}
This work was supported by the National Science Foundation grant: \#1755850, \#1841520, \#1907805, Jeffress Trust Award, and NVIDIA GPU Grant.

\bibliographystyle{ACM-Reference-Format}
\bibliography{cikm.bib}

\end{document}